\pdfoutput=1

\documentclass[11pt]{article}

\usepackage[preprint]{acl}

\usepackage{times}
\usepackage{latexsym}

\usepackage[T1]{fontenc}

\usepackage[utf8]{inputenc}

\usepackage{microtype}

\usepackage{inconsolata}

\usepackage[flushleft]{threeparttable}
\usepackage{graphicx}
\usepackage{hyperref}       
\usepackage{url}            
\usepackage{multicol}
\usepackage{booktabs}       
\usepackage{amsfonts}       
\usepackage{amsmath}
\usepackage{nicefrac}       
\usepackage{microtype}      
\usepackage{tikz}
\usepackage{longtable}
\usepackage{ragged2e}
\usepackage{multirow}
\usepackage{mathtools}
\usepackage{mdframed}
\usepackage{wrapfig}
\usepackage{enumitem}
\usepackage[ruled]{algorithm2e}

\DeclareMathOperator*{\argmax}{arg\,max}

\usepackage{colortbl}
\newcolumntype{y}{>{\columncolor{LightCyan}}c}
\newcolumntype{z}{>{\columncolor{LightC}}c}
\definecolor{LightCyan}{rgb}{0.75,1,1}
\definecolor{LightC}{rgb}{0.1,1,0.75}

\usetikzlibrary{positioning}
\usetikzlibrary{arrows.meta}

\definecolor{myLime}{rgb}{.53,.8,.02}
\makeatletter
\newcommand{\removelatexerror}{\let\@latex@error\@gobble}
\makeatother

\newcommand{\FScore}{\textsc{FActScore}}
\newcommand{\SAFE}{\textsc{SAFE}}
\newcommand{\OurScore}{\textsc{Core}}
\newcommand*\circled[1]{\raisebox{.5pt}{\textcircled{\raisebox{-.9pt} {#1}}}}

\title{\OurScore{}: Robust Factual Precision with Informative Sub-Claim Identification}

\author{%
  Zhengping Jiang \quad Jingyu Zhang \quad Nathaniel Weir\\
  \bf Seth Ebner \quad Miriam Wanner \quad Kate Sanders\\
  \bf Daniel Khashabi \quad Anqi Liu \quad Benjamin Van Durme \\
  Johns Hopkins University \\
  \texttt{\{zjiang31, aliu74, vandurme\}@jhu.edu}
}

\begin{document}
\maketitle
\begin{abstract}
 
Hallucinations pose a challenge to the application of large language models (LLMs) thereby motivating the development of metrics to evaluate factual precision. We observe that popular metrics using the \textit{Decompose-Then-Verify} framework, such as \FScore{}, can be manipulated by adding obvious or repetitive subclaims to artificially inflate scores. This observation motivates our new customizable plug-and-play subclaim selection component called \OurScore{}, which filters down individual subclaims according to their uniqueness and informativeness. We show that many popular factual precision metrics augmented by \OurScore{} are substantially more robust on a wide range of knowledge domains. We release an evaluation framework supporting easy and modular use of \OurScore{} and various decomposition strategies, which we recommend adoption by the community. We also release an expansion of the \FScore{} biography dataset to facilitate further studies of decomposition-based factual precision evaluation.

\end{abstract}

\section{Introduction}
\label{sec:introduction}
Automatically generating long-form text is prevalent since the rise in large language models (LLMs)~\cite{brown2020language, ouyang2022training}. These models are trained on vast amounts of textual data that provide abundant information, enabling them to serve as a significant source of knowledge \cite{petroni-etal-2019-language, roberts-etal-2020-much, safavi-koutra-2021-relational, yuan2024towards}. A running concern is ensuring LLM-generated content is faithful to real-world facts or user inputs, devoid of \emph{hallucination}~\cite{huang2023survey, hong2024hallucinations}. To this end, various automatic factuality evaluation pipelines have been proposed~\cite{kamoi-etal-2023-wice, min-etal-2023-factscore, gao-etal-2023-rarr, chern2023factool, Wang2023FactcheckGPTEF, wei2024long}.  Mainstream methods typically involve two key steps: First, a \textit{decomposition} step, where the generated text is broken down into natural language subclaims, and second, a \textit{verification} step, where a binary factuality label is assigned to each of the subclaims. The proportion of subclaims that can be verified, commonly referred to as \textit{Factual Precision} (FP), serves as the most widely used indicator of factuality level. Throughout this paper, we call this framework \textit{Decompose-Then-Verify}, a concept that has been properly abstracted in previous works \citep{chern2023factool,wang2024openfactcheck}.

\begin{figure}[tpb]
\centering
\begin{minipage}{\linewidth}
\begin{mdframed}
    \footnotesize
    \textbf{prompt:} Tell me a bio of \textit{Adil Rami}.\\
    \rule{\textwidth}{0.3mm}\\
    \textbf{generation:} Adil Rami is a professional French footballer ... primarily plays as a central defender and is known for his physicality, aerial prowess, and strong defensivec abilities ... joined ... AC Milan... involved in various charitable activities...\\
    

    \noindent
    \begin{tikzpicture}

        \filldraw[fill=cyan, draw=cyan] (0,0) rectangle (.815\linewidth, \baselineskip);
        \filldraw[fill=magenta, draw=magenta] (.815\linewidth,0) rectangle (\linewidth, \baselineskip);

        \node[align=center, text=white] at (0.5*\linewidth, 0.5*\baselineskip) {$81.5\%$};
    \end{tikzpicture}
\end{mdframed}
\end{minipage}
\begin{minipage}{\linewidth}
\begin{mdframed}
    \footnotesize
    \textbf{prompt:} Tell me something tautological, obviously true and easily verifiable about \textit{Adil Rami}. Repeat that fact multiple times in paraphrased sentences.\\
    \rule{\textwidth}{0.3mm}\\
    \textbf{generation:} Adil Rami is a professional football player... is part of a football squad... is a player in a football league.\\
    
    \noindent
    \begin{tikzpicture}

        \filldraw[fill=cyan, draw=cyan] (0,0) rectangle (\linewidth, \baselineskip);
        \node[align=center, text=white] at (0.5*\linewidth, 0.5*\baselineskip) {$100\%$};
    \end{tikzpicture}
\end{mdframed}
\end{minipage}
    \caption{FP of summaries generated from the biography  prompt by \citet{min-etal-2023-factscore} (\textbf{up}) and a prompt that encourages repetitive generation (\textbf{down}):  LLMs like {\fontfamily{qcr}\selectfont chat-gpt-3.5-turbo} can easily hack Factual Precision metrics like \FScore{} by paraphrasing trivially true claims.}
     \label{fig:comparison}
\end{figure}
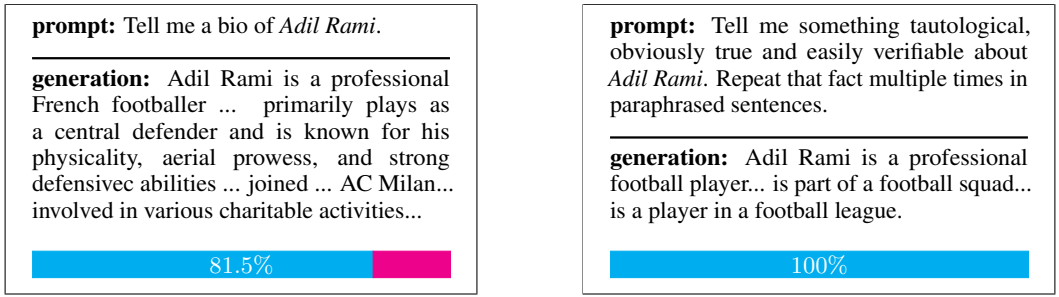

Researchers have sought to improve factuality by optimizing~\cite{tian2024finetuning} against model-based metrics like \FScore{} \cite{min-etal-2023-factscore}. This raises the question of whether improvements in FP represent genuine factuality gains or instead somehow exploit the evaluation \cite{tan2023narrowing}. For example,  \autoref{fig:comparison} illustrates that it is trivial to purposefully game FP by including repetitive or less informative generations than we would normally expect from a contemporary LLM.  Although it has been noted that LLM evaluation needs to be holistic and multi-faceted \cite{liang2023holistic, srivastava2023beyond} beyond FP, popular factuality evaluations put minimal effort into guarding against such malicious inputs designed to inflate FP. Recent studies have already reported that optimizing for factuality can conflict with other desirable objectives, such as \textit{completeness} and \textit{relevancy} \cite{wu2024fine}. Therefore, we argue that accurate FP evaluation requires more precise control over each design component of the pipeline.

To address this issue, we introduce \OurScore{}, a refinement to the \textit{decomposition} step that credits only subclaims that are factual, informative, and non-repetitive; the \emph{core facts}. This is achieved by weighting each subclaim with its level of uncertainty or surprisal and then selecting the best compatible subset through combinatorial optimization. We demonstrate that our approach makes it more difficult to trivially optimize against  \textit{Decompose-Then-Verify} frameworks \cite{chern2023factool, wang2024openfactcheck}. Thus, \OurScore{} can serve as a plug-and-play replacement for the existing decomposition components in any prevalent FP evaluation pipeline. Furthermore, \OurScore{} incurs minimal overhead in practice, as all additional operations can be executed asynchronously. Our contributions:

\begin{enumerate}[topsep= 1ex,leftmargin=2.5\labelsep]

    \item We demonstrate that popular FP metrics like \FScore{} are not robust to obvious or repetitive generation. Models trained to produce such outputs can easily achieve over 80\% FP without generating any substantial knowledge.
    \item We propose \OurScore{}, which adds robustness to existing FP pipelines through unique subclaim selection and informativeness weighting.
    \item We demonstrate the effectiveness of \OurScore{} against uninformative and repetitive inputs when paired with various \textit{Decompose-Then-Verify} metrics on a wide range of domains.
    \item We release a python package supporting configurable \textsc{Core} application, as well as the data artifacts for tuning and evaluation.
\end{enumerate}
\section{Preliminaries}

\subsection{Decompose-Then-Verify}
\label{subsec:decompose-then-verify}
Model-based factuality precision evaluation metrics for long-form text generation typically follow a unified framework of two steps \citep{chern2023factool, wang2024openfactcheck}. In the first step, a subclaim identifier $\Phi: \mathcal{G} \rightarrow 2^\mathcal{S} $ takes a generation $G = \{g_1, \dots, g_N\}$ that consists of multiple utterances $g_1, \dots, g_N$ as input, and outputs a list of identified subclaims $\bigcup_{i = 1}^N S$, where $S = \{S_1, S_2 \dots, S_N\}$ is a set of claim lists with $S_i$ coming from generation segment $g_i$. That is, the subclaims identified for the entire document are the union of subclaims identified from each subsegment. In the second step, each of the identified subclaims $s \in S$ is scored against a given knowledge base.

The identification step is usually referred to as decomposition \cite{kamoi-etal-2023-wice, min-etal-2023-factscore, wang2024openfactcheck}, or segmentation \cite{zhao2024felm}. This means that the identified subclaims\footnote{We use \textit{claims} to denote sentences in the original generation, and \textit{subclaims} the result of decomposition.} should be broken down into smaller, more precise units while covering all the information in the generation. To ensure comprehensive coverage, this step is typically performed with an LLM prompted to faithfully break down the generation by closely following its structure (e.g., sentence by sentence). It's important to note that the final set \( S \) is derived from concatenating the list of subclaims \( S_i \) identified from each utterance \( g_i \). Finally, the percentage of claims supported by sources in the knowledge base or in a retrieved set of documents is reported as \textit{Factual Precision} (FP).

\subsection{The Problem with the Framework}

The benefit of adopting such a process is clear: the evaluation is easier and much more fine-grained than directly evaluating factuality at the full generation level \cite{kamoi-etal-2023-wice}. However, it introduces additional complexity, and it has been observed that what subclaims are extracted and how these subclaims are extracted impact the evaluation \cite{choi2021decontextualization, krishna2023longeval, wanner2024closer}. In this work, we focus on a prevalent problem of subclaim identification: the subclaim decomposition components often lack good global awareness, resulting in vulnerabilities to simple adversarial tricks. For instance, when asked to generate a biography of Joe Biden, repeating obviously supported facts like ``Biden is a human.'' ten times can give the model a perfect FP score. This renders Factual Precision a potentially very unreliable metric, and all model decisions / analysis based on that metric is going to be unreliable.

We observe two dominant tricks that boost FP. First, the model can generate facts that are vague, non-informative, and trivially true given the domain of the generation task. Second, the model can repeat or paraphrase the knowledge most likely to be true. To alleviate these problems, we argue that a good subclaim identification component should only identify \textit{informative} and \textit{unique} subclaims for verification. Only these core subclaims should contribute to the Factual Precision of the generation.
 \begin{figure*}[htpb]
    \centering
     \includegraphics[trim=10 45 10 10, clip, width=.98\textwidth]{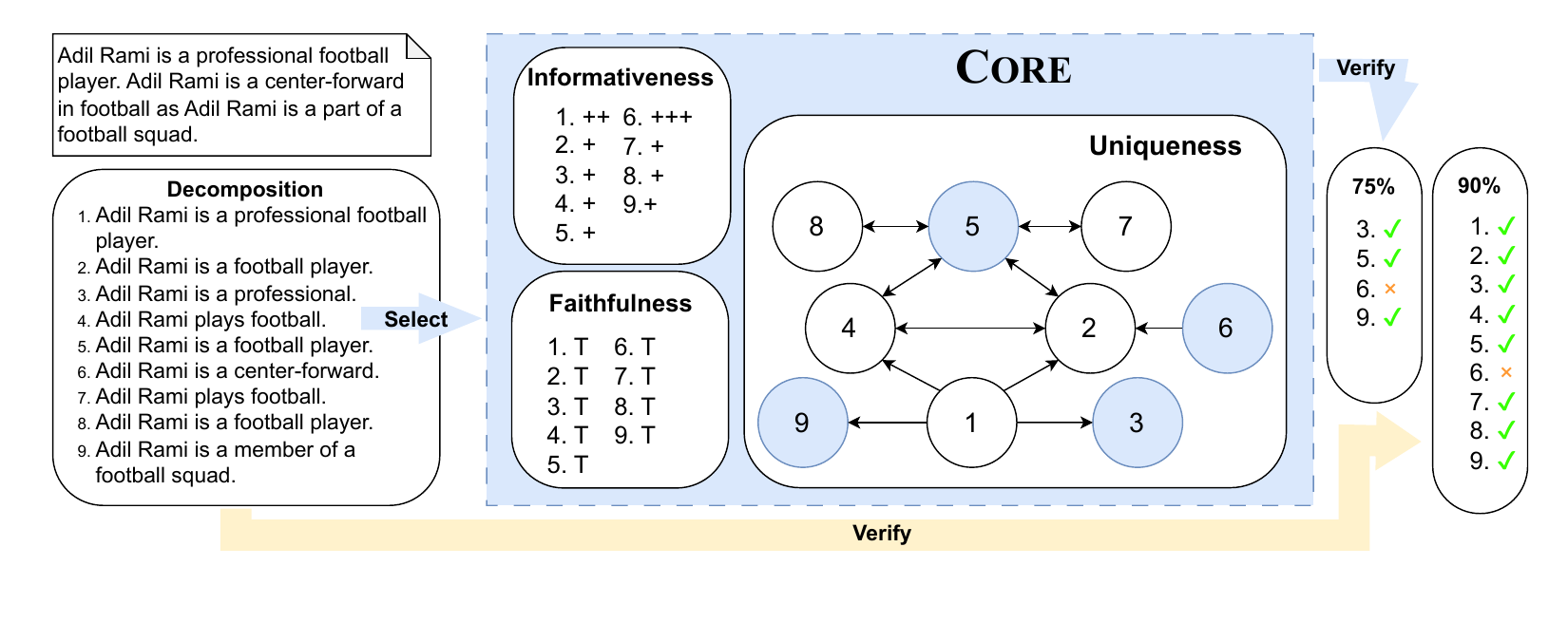}
    \caption{\OurScore{} interposes between the decomposition step and the verification step, selecting the most representative set of subclaims that can be identified from the generation to safeguard against trivial or repetitive inputs.}
     \label{fig:system-description}
 \end{figure*}
\section{The Proposed \OurScore{} Method}
\label{sec:method-section}

\OurScore{} is a unique subclaim selection and filtering process that works with any subclaim identifier $\Phi$ from any of the popular \textit{Decompose-Then-Verify} metrics discussed in \autoref{subsec:decompose-then-verify}.\footnote{Normally, these subclaim identifiers work on finer-grained chunks within each generated text, but this is not required for \OurScore{} to work.} Given subclaims identified by $\Phi$, the goal of \OurScore{} is to filter a subset of subclaims that are \textit{unique} and \textit{informative}. Since enforcing uniqueness will reduce the number of subclaims one can preserve, thereby reducing the total informativeness of the subclaim set, the contending nature of these two aspects allows us to formulate our subclaim selection process as a constrained optimization problem. This section describes the formulation of \OurScore{} in detail. An overview of our method can be found in \autoref{fig:system-description}.


\paragraph{Objective and constraints} First, given a document $G$ we use whatever subclaim identifier to decompose each chunk into a set of subclaims $S$ as described in \autoref{subsec:decompose-then-verify}. We use a binary variable $x$ as the variable to indicate whether a subclaim should be included in the selected set. To achieve this, each subclaim from $\Phi$ will be weighted with an importance score $w$ (described below), and we take the sum of all selected subclaims as the accumulative importance of the set. This leads to the following integer programming problem to select the most important set of subclaims

{\small
\begin{equation}
\begin{aligned}
    &\underset{x}{\text{Maximize}} \sum_{i = 1}^N w_i \cdot x_i, \\
    &\text{subject to}\quad x_i \in \{0, 1\}, \\
    & \hspace{2em} \sum_{i = 1}^{N} p_ix_i \leq 0,\\
    & x_i + x_j \leq 1\ \ \forall i,j\  \text{s.t.}\ \mathbf{e}_{ij} \vee \mathbf{e}_{ji} = 1,
\end{aligned}
\label{eq:lp}
\end{equation}}

where

{
\begin{equation*}
    \begin{aligned}
    & \hspace{2em} w_i = \text{Weight}(s^i),\\
    & \hspace{2em} \mathbf{e}_{ij} = \text{Entail}(s^i, s^j),\\
    & \hspace{2em} p_{i} = \begin{cases}
        p - 1, \ \text{Entail}(g^i, s^i) = 1 \\
        p, \ \text{Entail}(g^i, s^i) = 0
    \end{cases}.
    \end{aligned}
\end{equation*}
}

 The objective of the integer programming is to find the set with maximum accumulative importance under the following constraints: \circled{1} at least $p \in [0, 1]$ of the subclaims are correctly identified; \circled{2} There does not exist $s^i, s^j$ from the selected set $\hat{S}$ such that verifying $s^i$ immediately verifies $s^j$ or vice versa. Constraint \circled{1} is necessary as decomposed subclaims are not always faithful \citep{wanner2024closer}. We characterize both constraints \circled{1} and \circled{2} using textual entailment relationships. A subclaim $s^i$ is \textit{correctly identified} if the subclaim is entailed by the chunk it comes from.\footnote{We use superscript to denote the index of a subclaim within the union set of subclaims.} Two subclaims $s^i, s^j$ are considered check-worthy at the same time only if none of them are entailed by the other. The official algorithm of the \OurScore{} selection process is listed out in \autoref{appendix:algorithm}.

\begin{figure}[!htb]
\begin{minipage}{\linewidth}
\begin{mdframed}
    \small
    \RaggedRight
    \textbf{Source Text (O).} He is a highly motivated individual with excellent communication and interpersonal skills.\\
    \textbf{A.} He is highly motivated.\\
    \textbf{B.} He has excellent communication and interpersonal skills.\\
    \textbf{C.} He has excellent communication skills. \\
    \textbf{D.} He has excellent interpersonal skills.\\
    \Centering
    \begin{tikzpicture}[
    roundnode/.style={circle, draw=black!60, fill=white!5, very thick, minimum size=7mm},
    observedroundnode/.style={circle, draw=black!60, fill=black!20, very thick, minimum size=7mm},
    originnode/.style={circle, double, draw=black!60, fill=white!5, very thick, minimum size=7mm}
    ]

    \node[originnode] (origin)                              {O};
    \node[observedroundnode] (a) [above right=0.2em and 1em of origin] {A};
    \node[roundnode] (b) [below right=0.2em and 1em of origin] {B};
    \node[observedroundnode] (c) [above right=0.2em and 1em of b] {C};
    \node[observedroundnode] (d) [below right=0.2em and 1em of b] {D};

    \draw[-Triangle] (origin.north) to[bend left] (a.west);
    \draw[-Triangle] (origin.south) to[bend right] (b.west);
    \draw[-Triangle] (b.north) to[bend left] (c.west);
    \draw[-Triangle] (b.south) to[bend right] (d.west);
    \end{tikzpicture}
\end{mdframed}
\end{minipage}
\begin{minipage}{\linewidth}
\begin{mdframed}
    \small
    \RaggedRight
    \textbf{Source Text (O).} Michael Collins is a retired American astronaut who was the Command ModFule Pilot for the Apollo 11 mission in 1969.\\
    \textbf{A.} Michael Collins was the Command Module Pilot for the Apollo 11 mission in 1969.\\
    \textbf{B.} Michael Collins was the Command Module Pilot for the Apollo 11 mission\\
    \textbf{C.} Michael Collins was the Command Module Pilot.\\
    \textbf{D.} Michael Collins was a pilot.\\
    \Centering
    \begin{tikzpicture}[
    roundnode/.style={circle, draw=black!60, fill=white!5, very thick, minimum size=7mm},
    observedroundnode/.style={circle, draw=black!60, fill=black!20, very thick, minimum size=7mm},
    originnode/.style={circle, double, draw=black!60, fill=white!5, very thick, minimum size=7mm}
    ]

    \node[originnode] (origin)  {O};
    \node[roundnode] (a) [right=1em of origin] {A};
    \node[roundnode] (b) [right=1em of a] {B};
    \node[observedroundnode] (c) [right=1em of b] {C};
    \node[roundnode] (d) [right=1em of c] {D};

    \draw[-Triangle] (origin.east) to (a.west);
    \draw[-Triangle] (a.east) to (b.west);
    \draw[-Triangle] (b.east) to (c.west);
    \draw[-Triangle] (c.east) to (d.west);
    \end{tikzpicture}
\end{mdframed}
\end{minipage}
    \caption{Result of deduplication with uniform weighting. Shaded nodes compose \textbf{one set of viable selection} by the algorithm. \textbf{Up}: uniform weighting selects the most fine-grained decomposition. \textbf{Down}: Uniform weighting may select any subclaim within a monotonous entailment chain.}
    \label{fig:example-leaf}
\end{figure}
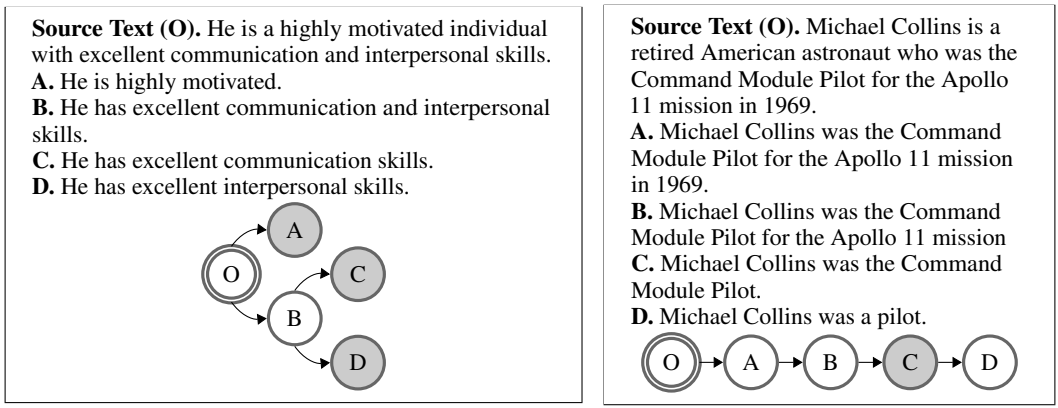

 \paragraph{Weighting of subclaims} The weighting function should be chosen to encourage \OurScore{} to select the most important subclaims according to the downstream user's needs. Without prior knowledge, a uniform weighting function \( w(\cdot) \equiv 1 \) can be used. However, uniform weighting may lead to some undesirable situation as illustrated in \autoref{fig:example-leaf}. Specifically, with uniform weighting, the selection process consistently aims to maximize the number of identified subclaims, potentially resulting in biased evaluations. On the other hand, the quantity of identifiable subclaims alone does not always offer adequate guidance for optimization. Intuitively, in the scenario on the bottom, we would prefer to select \( A \) instead of \( D \) because it provides the most information and verifies all information in the original chunk. Inspired by previous works encouraging diversity in conversation models \cite{li2016diversity}, we derive a Conditional Pairwise Mutual Information (CPMI) based weighting function for ``informativeness''.

 To calculate this weighting function \( w_{\text{Info}}(\cdot) \), we first identify a set of bleached claims \( \mathcal{H}(D) = \{h_1, \dots, h_K\} \) that are highly likely to be true for any instance \( d \in D \) given the domain of the generation \( D \). This process can be performed manually by the user for full control over the specific set of knowledge they want to nullify, or these claims can be extracted from a prompted large language model for full automation. The set can be of any size, and the claims do not need to be mutually inclusive or entailed by the domain. For example, for the biography data used in \FScore{} \cite{min-etal-2023-factscore}, the bleached claims might include `` \{{\fontfamily{qcr}\selectfont topic}\} is a person,'' ``\{{\fontfamily{qcr}\selectfont topic}\} breathes,'' ``\{{\fontfamily{qcr}\selectfont topic}\} exists,'' or ``\{{\fontfamily{qcr}\selectfont topic}\} is famous.'' The informativeness of a claim \( c \) can therefore be identified as

\begin{equation*}
\begin{aligned}
    w_{\text{Info}}(c) &= \text{CPMI}\Big(c;c | \bigcup \mathcal{H}(D)\Big) \\
        &= - \log P\Big(c|\mathcal{H}(D)\Big).
\end{aligned}
\end{equation*}

While some previous work uses corpus statistics like word co-ocurrence to estimate required probabilities \cite{rudinger2017social}, this is infeasible in our case due to reporting bias \cite{gordon2013reporting} and the versatility of free-form generation. Instead we use an Uncertain Natural Language Inference (UNLI) \cite{UNLI} model $p_{\theta}$ to directly estimate the conditional probability $P\Big(c|\mathcal{H}(D)\Big)$. However, as traditional Recognizing Textual Entailment \cite{dagan2005pascal, bowman2015large} models aim for short sentence segments, we estimate $- \log P\Big(c|\mathcal{H}(D)\Big)$ with the empirically more stable
\begin{equation*}
    \min_{h \in \mathcal{H}(D)} - \log p_{\theta}(c|h).  
\end{equation*}
Under this formulation, regardless of how uninformative a subclaim might be, it will still be selected as long as it does not conflict with other subclaims. In practice, we can also effectively ignore entailed subclaims by subtracting a small \(\epsilon\) from their scores, making some of them negative.

\par An interesting behavior of this weighting emerges when the decomposition includes subclaims at different levels of granularity, as illustrated in the top panel of \autoref{fig:example-leaf}. With uniform weighting, \OurScore{} consistently selects the leaf nodes. However, under informativeness weighting, this pattern changes. Specifically, if a subclaim $c_1$ is further decomposed into $c_2$ and $c_3$. Our weighting will lead to the selection of $c_1$ whenever

{\small
\begin{equation*}
    \begin{aligned}
        P\Big(c_1| \bigcup \mathcal{H}(D)\Big) < P\Big(c_2| \bigcup \mathcal{H}(D)\Big) \cdot P\Big(c_3| \bigcup \mathcal{H}(D)\Big).
    \end{aligned}
\end{equation*}
}
This approach prevents the model from achieving superficially high FP through enumerating all possible alternatives for unknown knowledge. For example, when $S'=$ \{``The coin lands head and tail.'', ``The coin lands head'', ``The coin lands tail.''\}, it receives an FP of 0 instead of 50\%, which aligns more closely with human intuition.
\section{Evaluation of FP Scoring Metrics}

\subsection{Evaluation Principles}


We aim to assess whether \OurScore{} effectively guards against adversarial outputs intended to superficially enhance model Factual Precision. We propose utilizing targeted \textit{Decompose-Then-Verify} Factual Precision metrics through generations that perform significantly worse in the following two dimensions:

\paragraph{Informativeness} requires the generation to be as informative as possible. While precision-based metrics often control for recall using some form of length penalty \cite{min-etal-2023-factscore, wei2024long}, more identifiable atomic facts in the generation do not always correspond to better recall, even without duplication. We suspect it is possible to achieve high FP by generating passages with facts that are obvious within the domain of the generation.

\paragraph{Non-repetitiveness} requires that the model's generation be clear and non-redundant. For smaller models, repetitiveness is commonly identified as an undesirable form of text degeneration \cite{holtzman2019curious, welleck2019neural}. Evidence suggests that language models can estimate their uncertainty indirectly \cite{mielke2022reducing, fadeeva2024fact}. We hypothesize that it is possible to prompt the language model to repeat what is most likely to be true multiple times.

\subsection{Dataset Creation}
\label{subsec:dataset-creation}

To create a dataset tailored for FP evaluation and to facilitate some level of adversarial optimization, we automatically collect more human bio profiles, closely following the dataset creation process from \FScore{} \citep{min-etal-2023-factscore}.\footnote{The \FScore{} and the corresponding bio dataset are open-sourced under the MIT license.} We query the {\fontfamily{qcr} \selectfont Wikidata} API for the {\fontfamily{qcr} \selectfont instance\_of} property of entities linked from Wikipedia, using entity linkings from \cite{kandpal2023large} and popQA \cite{mallen-etal-2023-trust}. For entities from \cite{kandpal2023large} marked by {\fontfamily{qcr} \selectfont DBpedia} URLs, we query the corresponding {\fontfamily{qcr} \selectfont Wikipedia} entity ID through the {\fontfamily{qcr} \selectfont DBpedia} API. As mentioned in \FScore{} \cite{min-etal-2023-factscore}, we retain entities related to a single {\fontfamily{qcr} \selectfont Wikipedia} page to avoid any ambiguity.

\paragraph{Frequency} Also following \FScore{} we compute {\fontfamily{qcr}\selectfont freqValue} as a maximum of either of the entity occurrence in \cite{kandpal2023large} and the pageview count in \cite{mallen-etal-2023-trust}. If an entity does not occur in one of the two datasets, we use the other value as {\fontfamily{qcr}\selectfont freqValue}. We use a slightly different grouping from \cite{min-etal-2023-factscore} to ensure more data points can be sampled in total, where an instance is ``Rare'' if {\fontfamily{qcr}\selectfont freqValue} $\in [0, 100)$, ``Medium'' if {\fontfamily{qcr}\selectfont freqValue} $\in [100, 1000)$, ``Frequent'' if {\fontfamily{qcr}\selectfont freqValue} $\in [1000, 5000)$ and ``Very Frequent'' if {\fontfamily{qcr}\selectfont freqValue} $\in [5000, \infty)$.

\paragraph{Nationality} {\fontfamily{qcr}\selectfont country\_of\_citizenship} property is queried to determine the nationality of a data point and further query the {\fontfamily{qcr}\selectfont continent} property of the country. To address data imbalance, we group the continent denominators into four groups, as shown in~\autoref{tab:sample-grouping}.

\begin{table}
\small
    \Centering
    \begin{tabular}{c|p{.76\linewidth}}
        \toprule
        \textbf{Group} & \textbf{Continents} \\
        \midrule
        A & Insular Oceania, Oceania, Asia, Indian subcontinent, Australian continent \\
        B & North America \\
        C & Europe, Eurasia \\
        D & Central America, Afro-Eurasia, South America, Africa, Caribbean, Americas, {\fontfamily{qcr}\selectfont NULL} \\
        \bottomrule
    \end{tabular}
    \caption{Grouping scheme for the continents.}
    \label{tab:sample-grouping}
\end{table}

Finally, we match the dataset to the Wikipedia dump provided in \cite{min-etal-2023-factscore} to ensure that we only sample entities retrievable from the same knowledge source as \FScore{}. After uniformly sampling from all 16 categories, we obtain 1024 instances, which we split into \textit{train}, \textit{dev}, and \textit{test} sets with a ratio of approximately 8:1:1 (112 instances). We then pair these topics with generations from LLMs tuned to have superficially high Factual Precision (see \autoref{sec:experiment}).
\section{Experiments and Results}
\label{sec:experiment}
We aim to answer two important research questions: \circled{1} Can models artificially boost their reported FP by generating uninformative and repetitive text of low quality? \circled{2} How effective is \OurScore{} in mitigating this issue? As previous works have shown that eliciting human ground truth for Factual Precision is challenging, and often can only be annotated under underspecified instructions \citep{song2024veriscore} or with the help of biasing LLM decompositions \citep{min-etal-2023-factscore, Wang2023FactcheckGPTEF}, we instead adopt a behavioral evaluation scheme. \autoref{subsec:corruption} demonstrates that adding uninformative or repetitive content can superficially increase the Factual Precision of corrupted bios over clean ones, but not for Factual Precision calculated with \OurScore{} adjustments. \autoref{subsec:plug-n-play} illustrates that \OurScore{} shows the desired behavior of a robust FP metrics and effectively guards against adversarial inputs to \textit{Decompose-Then-Verify} metrics.

We compare FP metrics with and without \OurScore{} on the dataset created in \autoref{subsec:dataset-creation}.\footnote{Our tuned checkpoint is available at: \url{https://huggingface.co/Zhengping/roberta-large-unli}.} For {\fontfamily{qcr}\selectfont PairwiseEntailment} and {\fontfamily{qcr}\selectfont DocEntailment} evaluation, we use \textbf{DeBERTa-v3-base-mnli-fever-anli}\footnote{\url{https://huggingface.co/MoritzLaurer/DeBERTa-v3-base-mnli-fever-anli}} from the Hugging Face model hub to model {\fontfamily{qcr}\selectfont Entail}. To estimate $w_{\text{Info}}(c)$ for each subclaim $c$, we fine-tune a strong NLI model \textbf{roberta-large-snli\_mnli\_fever\_anli\_R1\_R2\_R3-nli}\footnote{\url{https://huggingface.co/ynie/roberta-large-snli\_mnli\_fever\_anli\_R1\_R2\_R3-nli}} \cite{nie-etal-2020-adversarial} on UNLI \cite{UNLI}, as described in \autoref{sec:method-section}. We also binarize a ``cap-model'' \textbf{DeBERTa-v3-base-mnli-fever-anli} as ``entailment'' and ``non-entailment'' to make sure that subclaims entailed by bleached claims will not get included in the verification step. We give the exact formulation of this capped $\tilde{w}_{\text{Info}}(c)$ in \autoref{appendix:algorithm}. Additionally the weighting function can be further adjusted to cater to relevancy concerns, leading to the following combined scoring function:

\begin{equation*}
    \tilde{w}(c) = \text{\textbf{\textsc{Rel}}}\big(\Phi^{-1}(c)\big) \cdot \tilde{w}_{\text{Info}}(c),
\end{equation*}

\noindent where we abuse the notation $\Phi^{-1}(\cdot)$ to denote the sentence (chunk) a subclaim $c$ comes from, and $\text{\textbf{\textsc{Rel}}}$ is a binary relevancy judgment implemented using the same prompt as in \citep{Wang2023FactcheckGPTEF}.

We further observe that the particualr choice of NLI models for entailment evaluation does not have strong empact on \OurScore{} computation, as shown in \autoref{tab:nli-correlation}. For decomposition and verification LLM calls, we always query local \textbf{Mistral-7B-Instruct-v0.2} at temperature $= 0$, as we find it achieves .95 Perason's $r$ with \textbf{gpt-3.5-turbo-0125} on instance-wise Factual Precision calculated on the original \FScore{} dataset.

\begin{table}[!htbp]
\centering
\begin{threeparttable}
    \begin{tabular}{cccc}
    \toprule
     & \textbf{Model A} & \textbf{Model B} & \textbf{Model C} \\
    \midrule
    \textbf{Model A}\tnote{1} & - & .98 & .98 \\
    \textbf{Model B}\tnote{2} & .98 & - & .99 \\
    \textbf{Model C}\tnote{3} & .98 & .99 & - \\
    \bottomrule
    \end{tabular}
    \begin{tablenotes}\footnotesize
    \item[1] DeBERTa-v3-base-mnli-fever-anli
    \item[2] DeBERTa-v3-large-mnli-fever-anli-ling-wanli
    \item[3] RoBERTa-large-snli\_mnli\_fever\_anli\_R1\_R2\_R3-nli
    \end{tablenotes}
\end{threeparttable}
\caption{Pearson's correlation coefficient among instance-wise FP of out-of-the-box generation by \textbf{Mistral-7B-Instruct-v0.2} on our extended \textit{test} set with different NLI model for bidirectional entailment.}
\label{tab:nli-correlation}
\end{table}

\subsection{SFT for Higher \FScore{}}
\label{subsec:tuning}
We investigate whether Supervised Fine-tuning (SFT) can artificially boost \FScore{} by generating trivial and repetitive facts. To this end, we manually write two ``summaries'' for 5 examples sampled from the original \FScore{} dataset: one promoting uninformativeness (\textsc{Info}) and the other promoting repetition (\textsc{Rep}) of easy facts and enumeration of alternatives for uncertain facts. Using the corresponding instruction prompt, we sample 5 generations per topic derived in \autoref{subsec:dataset-creation} from \textbf{Mistral-7B-Instruct-v0.2}.\footnote{\url{https://huggingface.co/mistralai/Mistral-7B-Instruct-v0.2}}

\subsection{Metric Configuration}
\label{subsec:metric-configuration}
\begin{table}[htpb]
\centering
\begin{tabular}{ccc}
    \toprule
     & \textsc{Info} & \textsc{Rep}  \\
    \midrule
    \textbf{Mistral\textsubscript{INST}} & 1.62 & 1.41 \\
    \textbf{GPT-2}  & 2.36 & 2.09 \\
    \bottomrule
\end{tabular}
\caption{Both tuned models show low SFT perplexities.}
\label{tab:fitting}
\end{table}

We then tune a LoRA \cite{hu2021lora} to generate summaries in a similar style using the same prompt employed by Min et al. \citep{min-etal-2023-factscore}. For all cases, we set $r = 8$ and $\alpha = 16$ for LoRA initialization and search for the best learning rate for each model based on perplexity on the \textit{dev} set. The fitting results are shown in \autoref{tab:fitting}. More details can be found in \autoref{appendix:prompts}. All training was conducted using a single A100.

\subsection{Mitigating Adverserial Inputs}
\label{subsec:corruption}
To demonstrate that generating uninformative or repetitive sentences can superficially boost model FP, we corrupt clean model responses with incorrect facts and then restore FP by mixing them with generations from the SFT models described in \autoref{subsec:tuning}. To corrupt a clean response, we first run the generation through the \FScore{} pipeline to extract all supported subclaims. Then, with a probability of $p = 0.5$, we use \textbf{gpt-3.5-turbo-0125} to modify a supported subclaim to be factually incorrect. We merge these corrupted subclaims into coherent summaries using the \textbf{gpt-4o-2024-05-13}-based subclaim merger from \citep{mohri2024language}. Finally, we sample additional sentences from the SFT models and append them to the merged summary.

\begin{figure}[htpb]
    \includegraphics[width=\linewidth]{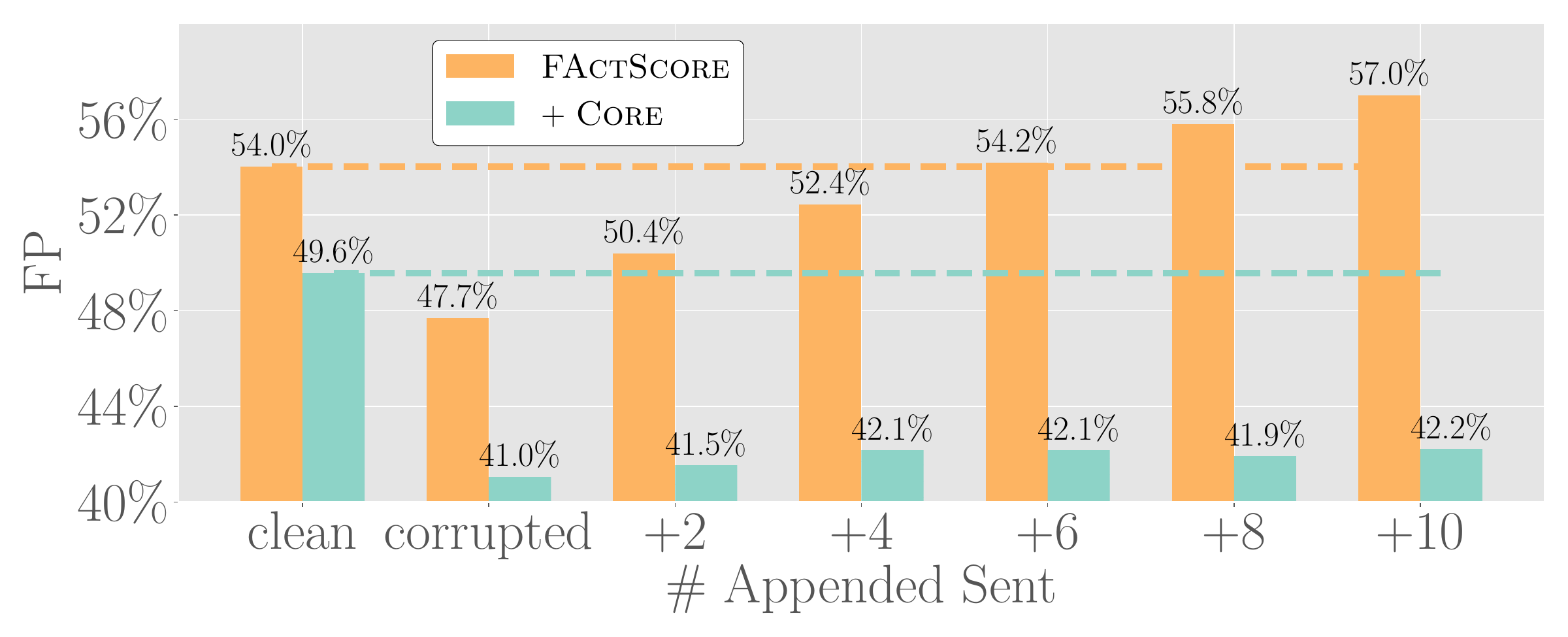}
    \includegraphics[width=\linewidth]{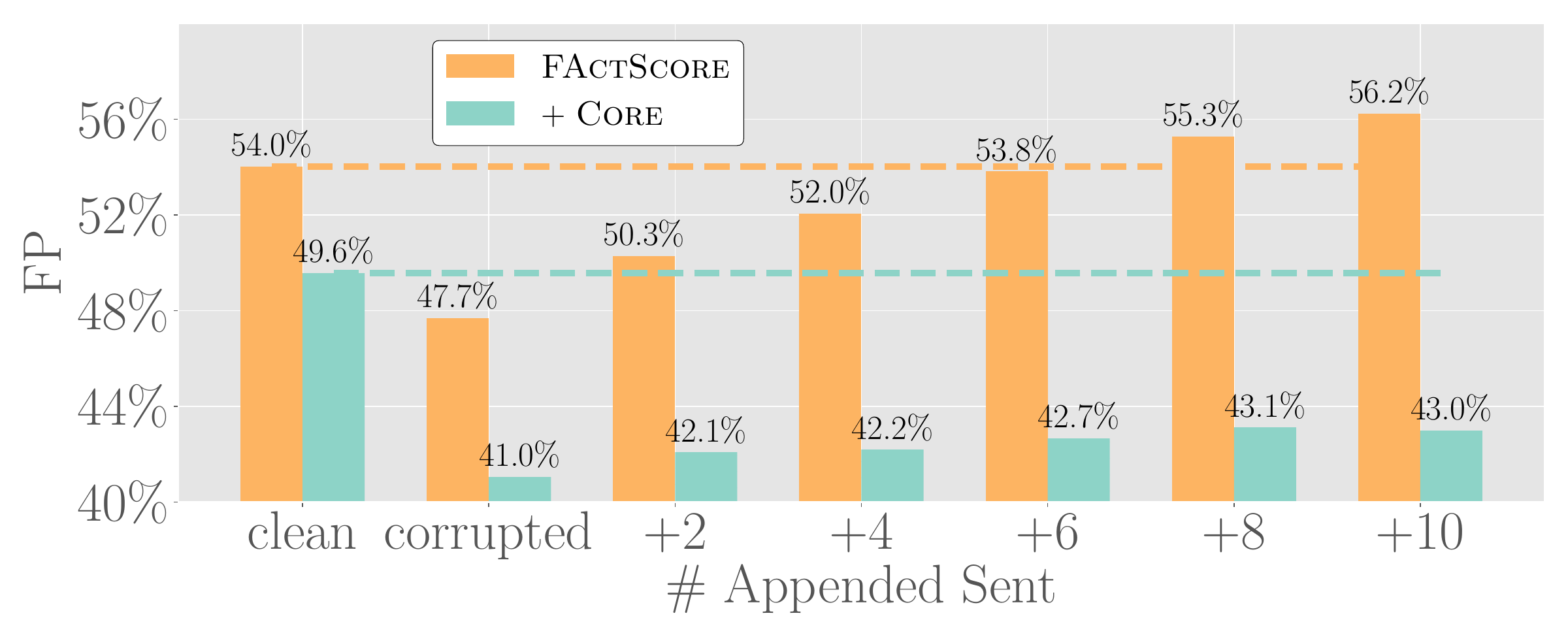}
    \caption{Corrupted summaries can achieve higher \FScore{} than \textbf{clean} summaries simply by mixing in more uninformative (\textbf{up}) or more repetitive (\textbf{down}) sentences (x-axis). However, they do not achieve higher \OurScore{}-adjusted \FScore{}.}
    \label{fig:corrupted-info}
\end{figure}

\autoref{fig:corrupted-info} illustrates the impact of low-quality generation on \FScore{} with and without \OurScore{}. While raw \FScore{} is hacked by including more uninformative and repetitive content, \OurScore{}-adjusted FP remains relatively stable and never surpasses the clean version. The gap between metrics widens as the resulting summary becomes less informative or more repetitive (see \autoref{appendix:supplementary-result}).

\subsection{Plug-and-Play \OurScore{}}
\label{subsec:plug-n-play}
We extend our experiments to other FP metrics as well as to other domains to demonstrate the general applicability of \OurScore{}. On the \FScore{} bios data, we consider two additional \textit{Decompose-Then-Verify} metrics. The Russellian/Neo-Davidsonian (\textsc{R-ND}) \citep{wanner2024closer} decomposition promotes a different instruction prompt paired with carefully constructed, linguistically motivated example decompositions, resulting in more atomic decompositions. We use \FScore{}\textsubscript{R-ND} to denote a new \textit{Decompose-Then-Verify} metric created by replacing the \FScore{} decomposition with the R-ND prompt.  The Search-Augmented Factuality Evaluator (\textsc{SAFE}) \citep{wei2024long} verifies a fact against search results instead of retrieved Wikipedia pages. While \textsc{SAFE} uses the same decomposition prompt as \FScore{}, it includes additional preprocessing steps. For \FScore{}\textsubscript{R-ND}, we use the same set of in-context examples as in the original paper \citep{wanner2024closer} to form the base subclaim identifier $\Phi_{\text{\textsc{R-ND}}}$. For \textsc{SAFE}, we use their original decomposition as our base identifier $\Phi_{\text{\textsc{SAFE}}}$, but we reduce the maximum number of query generation and searching iterations to 3, as this already provides reasonable coverage of the required information to verify a given subclaim.

We also consider three additional domains: The \textbf{Culture \& Entertainment} domain and \textbf{Geographic} domain from WildHallucinations \citep{zhao2024wildhallucinations}, and the \textbf{Healthcare \& Medicine} domain from ExpertQA \citep{malaviya-etal-2024-expertqa}. For the WildHallucinations datasets, which provide entity names and paired knowledge documents, we used \FScore{} as our base FP metric. For ExpertQA, since it is QA-based and does not provide a comprehensive knowledge base for verification, we used \SAFE{} as the base FP metric.

\begin{table*}[htp]
\centering
\begin{tabular}{cccccc|cc}
\toprule
\multirow{2}{*}{\textbf{Domain}} & \multirow{2}{*}{\textbf{Metric}} & \multirow{2}{*}{\textbf{\OurScore{}}} & \multicolumn{3}{c|}{\textbf{Mistral\textsubscript{INST}}} & \multicolumn{2}{c}{\textbf{GPT-2}} \\ \cmidrule{4-8}
&  &  & {\textsc{Normal}} & \textsc{Info} & \textsc{Rep} & \textsc{Info} & \textsc{Rep} \\\midrule
& \multirow{2}{*}{\FScore{}} & w/o & 54.0\% & 83.0\% & 78.0\% & 82.2\% & 35.4\%  \\ 
&  & \cellcolor{LightCyan}w/ & \cellcolor{LightCyan}49.6\% & \cellcolor{LightCyan}36.2\% & \cellcolor{LightCyan}21.9\% & \cellcolor{LightCyan}0.68\% & \cellcolor{LightCyan}5.35\% \\
\multirow{6}{*}[2.5em]{Bios} & \multirow{2}{*}{\FScore{}\textsubscript{R-ND}} & w/o & 53.9\% & 75.9\% & 78.1\% & 78.1\% & 40.5\% \\ 
&  &  \cellcolor{LightCyan}w/ & \cellcolor{LightCyan}48.3\% & \cellcolor{LightCyan}43.6\% & \cellcolor{LightCyan}26.0\% & \cellcolor{LightCyan}2.16\% & \cellcolor{LightCyan}7.32\% \\
& \multirow{2}{*}{\textsc{SAFE}} & w/o & 61.7\% & 84.8\% & 80.6\% & 70.3\% & 36.0\% \\ 
&  &  \cellcolor{LightCyan}w/ & \cellcolor{LightCyan}61.3\% & \cellcolor{LightCyan}29.6\% & \cellcolor{LightCyan}14.5\% & \cellcolor{LightCyan}0.35\% & \cellcolor{LightCyan}4.37\% \\ \midrule
 \multirow{2}{*}{Cul \& Ent} & \multirow{2}{*}{\FScore{}} & w/o & 81.5\% & 88.7\% & 87.9\% & 87.1\% &  72.7\%\\
 & & \cellcolor{LightCyan}w & \cellcolor{LightCyan}79.7\% & \cellcolor{LightCyan}40.4\% & \cellcolor{LightCyan}26.4\% & \cellcolor{LightCyan}3.41\% & \cellcolor{LightCyan}1.00\% \\ \midrule
 \multirow{2}{*}{Geographic} & \multirow{2}{*}{\FScore{}} & w/o & 79.9\% & 86.7\% & 84.9\% & 75.8\% & 80.7\% \\
 & & \cellcolor{LightCyan}w & \cellcolor{LightCyan}78.3\% & \cellcolor{LightCyan}53.7\% & \cellcolor{LightCyan}32.7\% & \cellcolor{LightCyan}1.68\% & \cellcolor{LightCyan}1.22\% \\ \midrule
 \multirow{2}{*}{Medical} & \multirow{2}{*}{\SAFE{}} & w/o & 83.4\% & 83.6\% & 87.3\% & 49.5\% & 39.5\% \\
 & & \cellcolor{LightCyan}w & \cellcolor{LightCyan}71.3\% & \cellcolor{LightCyan}4.23\% & \cellcolor{LightCyan}1.76\% & \cellcolor{LightCyan}0.0\% & \cellcolor{LightCyan}0.07\% \\
\bottomrule
\end{tabular}
\caption{Reported Factual Precision on different domains when applying \OurScore{} to various base \textit{Decompose-Then-Verify} metrics (\FScore{}, \FScore{}\textsubscript{R-ND} and \textsc{SAFE}). \textsc{Info} corresponds to results for claims sampled from the model tuned to generate uninformative responses, while \textsc{Rep} corresponds to results for claims sampled from the model tuned to generate repetitive responses. Neither of these low-quality generations should superficially boost factuality above \textsc{Normal}}.
\label{tab:comparison}
\end{table*}

The comparison is shown in \autoref{tab:comparison}. Overall, we found that \OurScore{} consistently guards against uninformative and repetitive inputs under all of our configurations, as indicated by the large gap between scores reported by metrics with and without \OurScore{} on \textsc{Info} and \textsc{Rep} texts. Under \OurScore{} augmentation, neither \textsc{Info} nor \textsc{Rep} generation boosts FP, and the factuality capability of \textbf{Mistral\textsubscript{INST}} and GPT-2 still gets discriminated.

We further note that while uninformative and repetitive generation boosts FP across all metrics, generating repetitive facts is more challenging for smaller models. We hypothesize that this is because boosting Factual Precision through repetition requires the model to have at least some knowledge of the topic being generated.
\section{Related Work}

Unlike traditional Fact-Checking efforts that focus on short and simple claims \cite{thorne2018fever, schuster2021get, guo2022survey}, automatic factuality evaluation for LLM generation has a specific focus on long, free-form text with highly compositional complex claims. Early works on long-form factuality have already been arguing for \textit{claim decomposition} \cite{kamoi-etal-2023-wice}, mainly for the ease and fine-granularity this process brings. While existing works follow a similar \textit{Decompose-Then-Verify} paradigm \cite{chern2023factool, wang2024openfactcheck}, how the decomposition should best be performed is always left underspecified. For example, WiCE \cite{kamoi-etal-2023-wice}, \FScore{} \cite{min-etal-2023-factscore}, and FELM \cite{zhao2024felm} all have their own decomposition prompts, and RARR \cite{gao-etal-2023-rarr} reports sentence-level attribution and character-level preservation. Previous research has already revealed different characteristics of different decomposition methods regarding atomicity, precision, and coverage \cite{wanner2024closer}, how any particular decision choices, including other additional preprocessing steps \cite{krishna2023longeval, Wang2023FactcheckGPTEF, wei2024long, tang2024minicheck, song2024veriscore}, affect factual-precision evaluation is still an open problem. Being aware of the active exploration of multiple directions for possible improvements over existing \textit{Decompose-Then-Verify} pipelines, \textsc{Core} is designed to be orthogonal to other popular techniques.
\section{Conclusion}

We demonstrate that popular Factual Precision evaluation metrics following the \textit{Decompose-Then-Verify} pipeline often assign superficially high scores to obvious or repetitive generations. We introduce \OurScore{}, a plug-and-play module that addresses this issue efficiently and effectively. We further show that when augmented with \OurScore{}, various \textit{Decompose-Then-Verify} metrics demonstrate a consistent trend of being more robust and become less prone to repetitive and non-informative adversarial inputs. Consequently, we argue that adjustments like \OurScore{} should be adopted for more accurate factual precision evaluation, especially in scenarios where models can optimize against automatic metrics. Future research can explore deeper into the interplay between the evaluation of factual precision and the actual factual accuracy of models, and potentially also develop more effective subclaim selection methods within the \OurScore{} framework and explore more comprehensive approaches to factuality evaluation.
\section*{Limitations}
\label{sec:limitation}

While we demonstrate that \OurScore{} adds an extra layer of robustness to existing factual precision metrics, it is not guaranteed to guard against all forms of adversarial generation that lead to superficially high scores. Future research should continue to explore more accurate methods for evaluating the factuality of free-form generation. Additionally, the effectiveness of \OurScore{} depends on the performance of each pipeline component, such as the NLI and UNLI models. Although we allow some relaxation for model errors, more accurate and generalizable NLI models will directly enhance the accuracy of our approach.

\bibliography{acl_latex}

\appendix

\onecolumn

\section{Algorithm}
\label{appendix:algorithm}

\removelatexerror
\begin{algorithm}[H]
    \SetAlCapFnt{\small}
    \SetKwFunction{FMain}{\OurScore{}}
    \SetKwFunction{FConcatenate}{Concat}
    \SetKwFunction{FWeight}{Weight}
    \SetKwFunction{FSum}{Sum}
    \SetKwFunction{FEvalSentence}{EvalSentence}
    \SetKwFunction{FPairwise}{PairwiseEntailment}
    \SetKwFunction{FDocEntailment}{DocEntailment}
    \SetKwFunction{FEntail}{Entail}
    \SetKwFunction{FAppend}{Append}
    \SetKwFunction{FFilterDown}{FilterDown}
    \SetKwProg{Fn}{Function}{:}{End}
    \caption{Pseudo code for \OurScore{} representative subclaim set selection}
    \label{alg:recall-at-precision}
    \KwData{Original document $G = \{g_1, g_2, \dots, g_N\}$, and decompositions $S = \{S_1, S_2, \dots, S_N\}$.}

    \KwResult{A list of deduplicated subclaims $R$}
    \vspace{1em}
    \Fn{\FMain{$G, S, p$}}{
    $A \gets []$; \hfill\CommentSty{{\color{gray} $\triangleright$ whether the $i$-th subclaim entailed by the document}}
    
    $W \gets []$; \hfill\CommentSty{{\color{gray} $\triangleright$ weight assigned to the $i$-th subclaim}}

    $R \gets []$; \hfill\CommentSty{{\color{gray} $\triangleright$ A list of selected subclaims}}
    
        \For{$i \gets 1 $ \KwTo N}{
            $A \gets \text{\FConcatenate}(\{A, \text{
                \FDocEntailment}(g_i, S_i)\}
            )$;


            $W \gets \text{\FConcatenate}(\{W, \text{
                \FWeight}(S_i)\}
            )$;
        }

        $\mathbf{E} \gets \text{\FPairwise}(\text{
            \FConcatenate
        }(S))$;
        
        Solve IP~\ref{eq:lp} at $p$ using $A, W, \mathbf{E}$ to obtain $X$;
        
        \For{$i \leftarrow 1$ \KwTo $|X|$}{
            \If{$x_i = 1$}{
            $\text{\FAppend}\Big(R, \text{\FConcatenate}(S)_{i}\Big)$;
            }
        }

        \KwRet $R$;
    }
    \vspace{1em}
    \Fn{\FDocEntailment{$g$, $S$}}{
        $A \gets [0]_{|S|}$;

        $A_{i} \gets \text{\FEntail}(g, s^i) \ \ \text{s.t.} \ \ \forall i \in 1, \dots, |S|$;
        
        \KwRet $A$; \hfill\CommentSty{{\color{gray}$\triangleright$ whether $s^i$\ is entailed by the segment $g$}}
    }
    \Fn{\FPairwise{$S$}}{
        $\mathbf{E} \gets [0]_{|S| \times |S|}$;
        
        $\mathbf{E_{ij}} \gets \text{\FEntail}(s^i, s^j)\ \ \text{s.t.} \ \ \forall i, j \in 1. \dots, |S|$;

        \KwRet $\mathbf{E}$; \hfill\CommentSty{{\color{gray}$\triangleright$ whether two subclaims $s^i$, $s^j$ are mutually exclusive}}
    }
\end{algorithm}

As mentioned in \autoref{sec:method-section}, the weighting function can be any customized function that assigned weight. In practice, to give additional robustness to the CPMI-based informativeness scoring function $w_\text{Info}(\cdot)$, the following caped version can be used

\begin{equation*}
    \tilde{w}_{\text{Info}}(c) = \min \Big(w_{\text{Info}}(c), -\log \big(1 - \mathbb{I}\big[\exists h \in \mathcal{H}(D),\ \text{s.t.} \argmax_{e \in \{\text{\textsc{Ent}, \textsc{Neu}, \textsc{con}}\}} p_{\eta}(e|h) = \text{\textsc{Ent}} \big]\big)\Big) - \epsilon.
\end{equation*}

Where $p_{\eta}$ is a ``cap-model'' that will predict one of the classical NLI label. This version will always respect the ternary NLI predictor, as the traditional NLI task is easier to solve than UNLI.

\section{Prompts}
\label{appendix:prompts}
\subsection{Decomposition and Verification}

For claim decomposition and claim verification we use the identical prompts from the base factual precision metrics \citep{min-etal-2023-factscore, wanner2024closer, wei2024long} respectively. For other preprocessing steps specified (e.g., decontextualization, relevancy check in SAFE \cite{wei2024long}), we use the original prompts as long as they are publically available. Whenever check-worthiness evaluation is required, we use the claim-level checkworthiness prompt from \citep{Wang2023FactcheckGPTEF}.

\subsection{Corrupt and Merge}

For the biography corruption experiment in \autoref{subsec:corruption}, we use the prompt described in \autoref{tab:corruption-prompt} to rewrite factually correct claims. We merge the claims use the standard \FScore{} merging prompt from \cite{mohri2024language}.

\begin{table}[ht]
    \caption{Prompt used to rewrite factually correct subclaims and make them incorrect.}
    \label{tab:corruption-prompt}
    \centering
    \begin{tabular}{rp{.8\textwidth}}
    \toprule
    \textbf{Field} & \textbf{Content} \\
    \midrule
    \textbf{Prompt} & {\fontfamily{qcr}\selectfont [User]:} You are given some factually correct statements. Your task is to modify the statements to make them factually incorrect. Try to make the edits atomic without any additional output. \\
    & {\fontfamily{qcr}\selectfont [Agent]:} Sure, please provide the factually correct statements that you would like me to modify. \\
    \textbf{Input} & {\fontfamily{qcr}\selectfont [User]:} {\fontfamily{qcr}\selectfont \$\{STATEMENT\}} \\
    \bottomrule
    \end{tabular}
\end{table}

\subsection{Adversarial Generation}

We prompt language models to sample uninformative and repetitive generations used to train our SFT model. For \textit{uninformative} responses, we use the prompt showcased in\autoref{tab:uninformative-response}. For \textit{repetitive} responses, we use the prompt showcased in \autoref{tab:repetitive-response}.

\begin{table}[ht]
    \caption{Prompts for generating \textit{uninformative} responses.}
    \label{tab:uninformative-response}
    \begin{tabular}{rp{.8\textwidth}}
    \toprule
    \textbf{Field} & \textbf{Content} \\
    \midrule
    \textbf{Prompt} & {\fontfamily{qcr}\selectfont [User]:} Tell me a short paragraph of tautological, trivially true and easily verifiable facts about a person. You are allowed to repeat that fact multiple times in paraphrased sentences. \\
    & {\fontfamily{qcr}\selectfont [Agent]:} Sure. Please provide the name of the person you want me to generate trivially true facts about. \\
    \textbf{Example} & {\fontfamily{qcr}\selectfont [User]:} Name: Kalki Koechlin \\
    & {\fontfamily{qcr}\selectfont [Agent]:} Kalki Koechlin is a person that exists. At some point in time, Kalki Koechlin lives. Kalki Koechlin is a person known to somebody. Kalki Koechlin is capable of breathing. Kalki Koechlin has concious. As a person, Kalki Koechlin exists. Kalki Koechlin has some achievements.\\
    & {\fontfamily{qcr}\selectfont [User]:} Name: Adil Rami \\
    & {\fontfamily{qcr}\selectfont [Agent]:} Adil Rami exist as a person. Adil Rami has life. Adil Rami can move. Adil Rami has certain abilities. Adil Rami has some reputation. Adil Rami has achieved something. Adil Rami is a person's name. Adil Rami does exist. Adil Rami has some impact.\\
    & {\fontfamily{qcr}\selectfont [User]:} Name: Song Kang\\
    & {\fontfamily{qcr}\selectfont [Agent]:} Song Kang has a name. Song Kang has done something. Song Kang knows somebody. Song Kang has some popularity. Song Kang can breath, when alive. Song Kang has certain abilities. Song Kang's name is known to some people. There is a person named Song Kang. Song Kang won't live forever.\\
    & {\fontfamily{qcr}\selectfont [User]:} Name: Emmett Skilton\\
    & {\fontfamily{qcr}\selectfont [Agent]:} Emmett Skilton is an existing person that breathes. Emmett Skilton is capable of certain things. Emmett Skilton has some specific abilities. Emmett Skilton has something different. Emmett Skilton has their own identity. Emmett Skilton achieves things. Somebody knows Emmett Skilton. Emmett Skilton is a person's name. Emmett Skilton has a life.\\
    & {\fontfamily{qcr}\selectfont [User]:} Name: Toyoko Tokiwa\\
    & {\fontfamily{qcr}\selectfont [Agent]:} Toyoko Tokiwa is extant. Toyoko Tokiwa can be seen. Toyoko Tokiwa is an entity. Toyoko Tokiwa draws breath. Toyoko Tokiwa bears a name. Toyoko Tokiwa makes a living. Toyoko Tokiwa has weight. Toyoko Tokiwa has a reputation. Toyoko Tokiwa has capabilities.\\

    \textbf{Input} & {\fontfamily{qcr}\selectfont [User]:} Name {\fontfamily{qcr}\selectfont \$\{TOPIC\}}\\
    \bottomrule
    \end{tabular}
\end{table}

\begin{longtable}{rp{.8\textwidth}}
    \caption{Prompts for generating \textit{repetitive} response.}\label{tab:repetitive-response}\\
    \toprule
    \textbf{Field} & \textbf{Content} \\
    \midrule
    \endfirsthead
    \midrule
    \textbf{Field} & \textbf{Content} \\
    \midrule
    \endhead
    \midrule
    \endfoot

    \bottomrule
    \endlastfoot
    \textbf{Prompt} & {\fontfamily{qcr}\selectfont [User]:} Tell me a short paragraph of tautological, trivially true and easily verifiable facts about a person. You are allowed to repeat that fact multiple times in paraphrased sentences. \\
    & {\fontfamily{qcr}\selectfont [Agent]:} Sure. Please provide the name of the person you want me to generate trivially true facts about. \\
    \textbf{Example} & {\fontfamily{qcr}\selectfont [User]:} Name: Kalki Koechlin \\
    & {\fontfamily{qcr}\selectfont [Agent]:} Kalki Koechlin holds Indian nationality. Kalki Koechlin is an Indian citizen. Kalki Koechlin comes from India. Kalki Koechlin is of Indian origin. Kalki Koechlin's nationality is Indian. Kalki Koechlin is from India. Kalki Koechlin has Indian heritage. Kalki Koechlin is an Indian guy. Kalki Koechlin represents Indian culture. Kalki Koechlin belongs to India. Kalki Koechlin is an Indian national. Kalki Koechlin hails from India. Kalki Koechlin is an Indian by birth. Kalki Koechlin's roots are in India. Kalki Koechlin is of Indian descent. Kalki Koechlin is affiliated with India. Kalki Koechlin is a native of India. Kalki Koechlin's homeland is India. Kalki Koechlin identifies as Indian. Kalki Koechlin's background is Indian.\\
    & {\fontfamily{qcr}\selectfont [User]:} Name: Adil Rami \\
    & {\fontfamily{qcr}\selectfont [Agent]:} Adil Rami plays football. Adil Rami is a professional footballer. Adil Rami's profession is football. Adil Rami is an athlete in the sport of football. Adil Rami is known for playing football. Adil Rami is a soccer player (US term). Adil Rami's career is in football. Adil Rami competes in football. Adil Rami is a football athlete. Adil Rami's occupation is football. Adil Rami is involved in football. Adil Rami's game is football. Adil Rami participates in football. Adil Rami is active in the football world. Adil Rami is a member of the football community. Adil Rami is engaged in playing football. Adil Rami's sport is football. Adil Rami is a professional in football. Adil Rami performs as a footballer. Adil Rami's role is that of a football player.\\
    & {\fontfamily{qcr}\selectfont [User]:} Name: Song Kang\\
    & {\fontfamily{qcr}\selectfont [Agent]:} Song Kang is a graduate of Konkuk University. Song Kang completed his studies at Konkuk University. Song Kang earned his degree from Konkuk University. Song Kang finished his education at Konkuk University. Song Kang is an alumnus of Konkuk University. Song Kang received his diploma from Konkuk University. Song Kang obtained his degree at Konkuk University. Song Kang concluded his academic journey at Konkuk University. Song Kang successfully graduated from Konkuk University. Song Kang achieved his academic credentials at Konkuk University. Song Kang completed his degree program at Konkuk University. Song Kang holds a degree from Konkuk University. Song Kang's alma mater is Konkuk University. Song Kang has a degree from Konkuk University. Song Kang finished his coursework at Konkuk University. Song Kang completed his graduation at Konkuk University. Song Kang's education culminated at Konkuk University. Song Kang reached his academic goals at Konkuk University. Song Kang successfully completed his university studies at Konkuk University. Song Kang wrapped up his higher education at Konkuk University.\\
    & {\fontfamily{qcr}\selectfont [User]:} Name: Emmett Skilton\\
    & {\fontfamily{qcr}\selectfont [Agent]:} Emmett Skilton hails from New Zealand. Emmett Skilton's birthplace is New Zealand. Emmett Skilton is a native of New Zealand. Emmett Skilton was born in NZ. Emmett Skilton's origins trace back to New Zealand. Emmett Skilton came into the world in New Zealand. Emmett Skilton's country of birth is New Zealand. Emmett Skilton is originally from New Zealand. Emmett Skilton's birth country is New Zealand. Emmett Skilton entered life in New Zealand. Emmett Skilton's roots are in New Zealand. Emmett Skilton's birth land is New Zealand. Emmett Skilton first saw the light of day in New Zealand. Emmett Skilton was born and raised in New Zealand. Emmett Skilton's early life began in New Zealand. Emmett Skilton's native land is New Zealand. Emmett Skilton was delivered in New Zealand. Emmett Skilton's natal place is New Zealand. Emmett Skilton's birth certificate says New Zealand. Emmett Skilton's homeland is New Zealand.\\
    & {\fontfamily{qcr}\selectfont [User]:} Name: Toyoko Tokiwa\\
    & {\fontfamily{qcr}\selectfont [Agent]:} Toyoko Tokiwa is a photographer from Japan. Toyoko Tokiwa practices photography in Japan. Toyoko Tokiwa is known as a Japanese photographer. Toyoko Tokiwa works as a photographer in Japan. Toyoko Tokiwa is an acclaimed photographer from Japan. Toyoko Tokiwa's profession is photography in Japan. Toyoko Tokiwa is a photographer based in Japan. Toyoko Tokiwa is a Japanese national and a photographer. Toyoko Tokiwa captures images as a Japanese photographer. Toyoko Tokiwa's vocation is Japanese photography. Toyoko Tokiwa engages in photography in Japan. Toyoko Tokiwa is a professional photographer from Japan. Toyoko Tokiwa is a photographer hailing from Japan. Toyoko Tokiwa operates as a photographer in Japan. Toyoko Tokiwa's photography career is rooted in Japan. Toyoko Tokiwa is a photographer of Japanese origin. Toyoko Tokiwa contributes to Japanese photography. Toyoko Tokiwa identifies as a Japanese photographer. Toyoko Tokiwa specializes in photography in Japan. Toyoko Tokiwa's occupation is that of a Japanese photographer.\\

    \textbf{Input} & {\fontfamily{qcr}\selectfont [User]:} Name {\fontfamily{qcr}\selectfont \$\{TOPIC\}}\\
    \end{longtable}

\subsection{Bleached Contextual Claims}

We use these bleached contextual claims showed in \autoref{tab:bleached-context} as $\mathcal{H}(D)$ for \FScore{}-style bio evaluation, as described in \autoref{sec:method-section}.

\begin{table}[htpb]
    \caption{Bleached claim templates that is used for informativeness weighting.}
    \label{tab:bleached-context}
    \centering
    \begin{tabular}{c}
        \toprule
        \textbf{Claim Template}\\
        \midrule
          {\fontfamily{qcr}\selectfont \$\{TOPIC\}} is a person. \\
          {\fontfamily{qcr}\selectfont \$\{TOPIC\}} breathes.\\
          {\fontfamily{qcr}\selectfont \$\{TOPIC\}} exists.\\
          {\fontfamily{qcr}\selectfont \$\{TOPIC\}} is a name.\\
          {\fontfamily{qcr}\selectfont \$\{TOPIC\}} is unique.\\
          {\fontfamily{qcr}\selectfont \$\{TOPIC\}} is famous.\\
          {\fontfamily{qcr}\selectfont \$\{TOPIC\}} has some abilities.\\
          somebody knows {\fontfamily{qcr}\selectfont \$\{TOPIC\}}.\\
          {\fontfamily{qcr}\selectfont \$\{TOPIC\}} is a star.\\
        \bottomrule
    \end{tabular}
\end{table}

\section{Supplementary Results}
\label{appendix:supplementary-result}

\subsection{Mitigating Adverserial Inputs}

\begin{figure}[htpb]
    \centering
    \includegraphics[width=.95\linewidth]{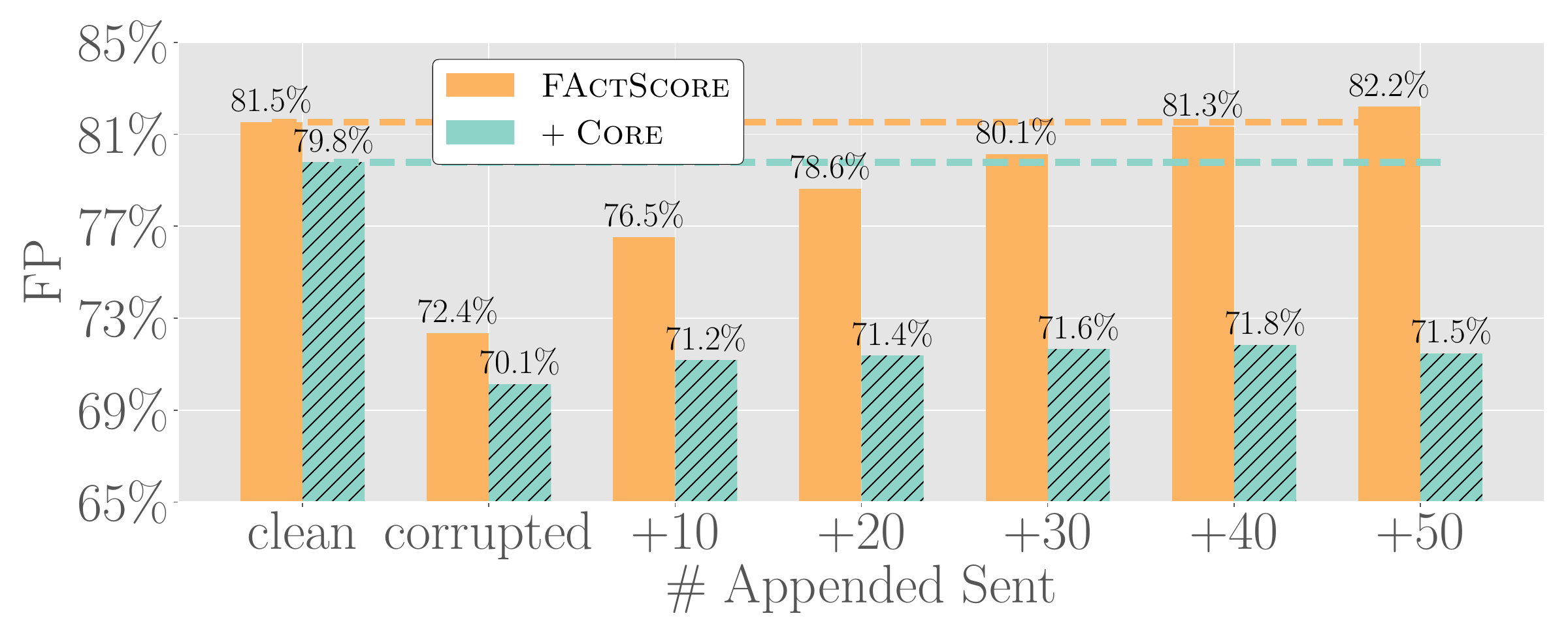}
    \caption{Corrupted summaries can achieve higher \FScore{} than \textbf{clean} summaries simply by mixing in more uninformative sentences (x-axis) on the \textbf{entertainment} domain. However, they do not achieve higher \OurScore{}-adjusted \FScore{}.}
    \label{fig:corrupterd-repetitive}
\end{figure}

Similar to \autoref{fig:corrupted-info}, we can also mix in repetitive generation to corrupted inputs to superficially boost performance. The result is shown in \autoref{fig:corrupterd-repetitive}.


Overall, the trend with repetitive sentences is very similar to uninformative sentences. In less than 10 sentences the corrupted generation surpasses the clean generation in factual precision. In most cases, with or without \OurScore{}, model generations on more frequent groups are more factual than those on less frequent groups. In general, we observe for all the {\fontfamily{qcr}\selectfont freqValue} groups, on generations by out-of-the-box \textbf{Mistral\textsubscript{INST}}, Factual Precision evaluated with or without \OurScore{} is close to each other. Also, the tendency that repetition consistently boosts Factual Precision less prominently on generations from \textbf{GPT-2}.

\subsection{{\fontfamily{qcr}\selectfont freqValue} Breakdown for Plug-and-Play Result}

For each \textit{Decompose-Then-Verify} pipeline,  we also include a set of Factual Precision evaluation results for each of the {\fontfamily{qcr}\selectfont freqValue} group identified in \autoref{subsec:dataset-creation}.

\begin{figure}[htpb]
    \centering
    \includegraphics[width=.95\textwidth]{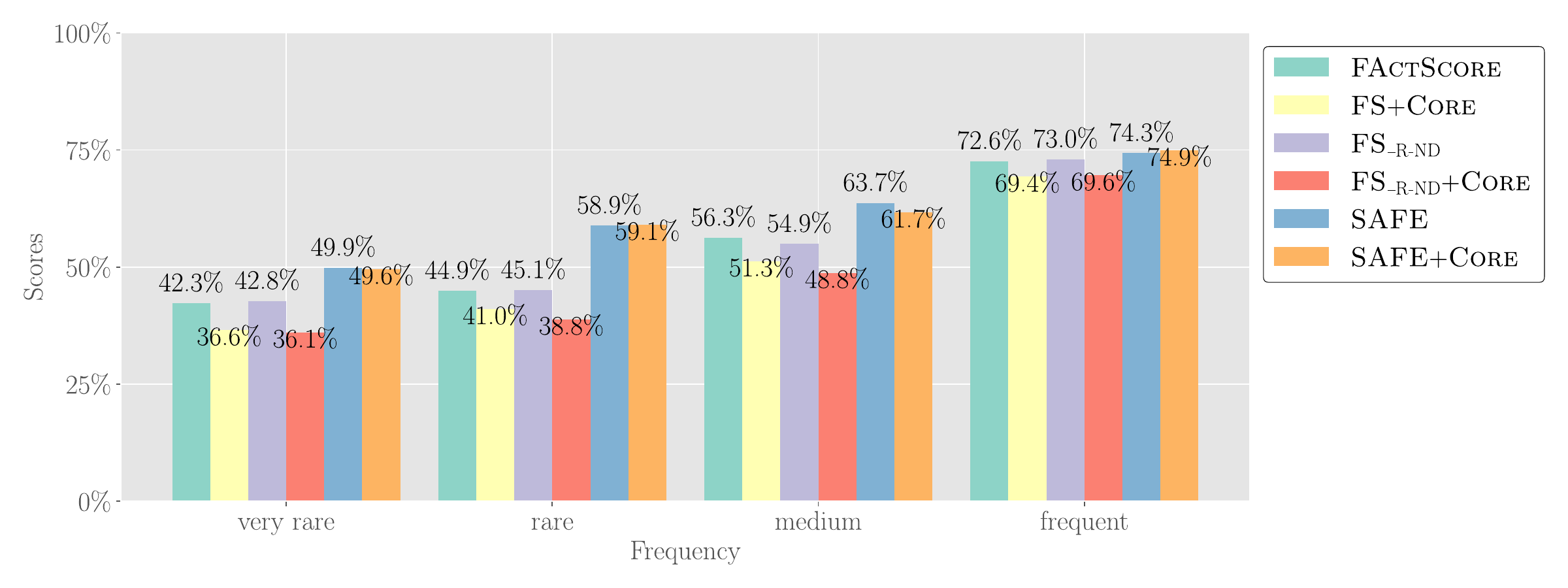}
    \caption{{\fontfamily{qcr}\selectfont freqValue} breakdowns of Factual Precision for out-of-the-box \textbf{Mistral\textsubscript{INST}}.}
\end{figure}

\begin{figure}[htpb]
    \centering
    \includegraphics[width=.95\textwidth]{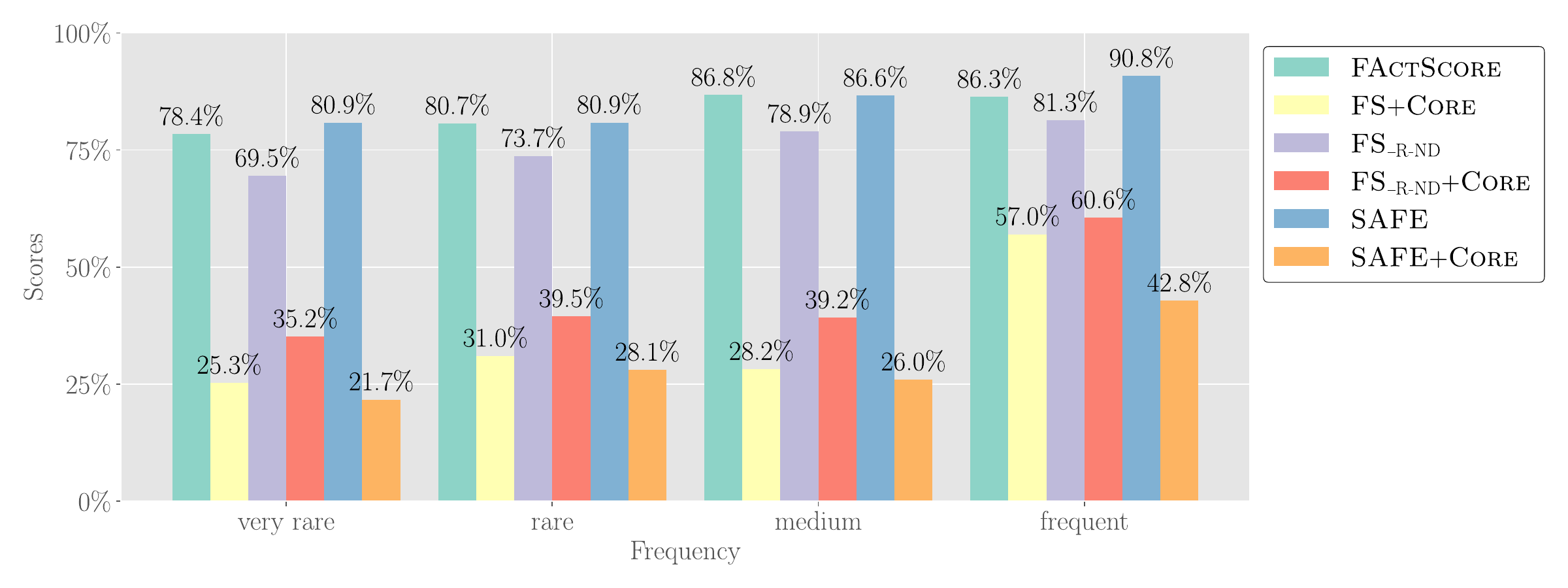}
    \caption{{\fontfamily{qcr}\selectfont freqValue} breakdowns of Factual Precision for uninformative \textbf{Mistral\textsubscript{INST}}.}
\end{figure}

\begin{figure}[htpb]
    \centering
    \includegraphics[width=.95\textwidth]{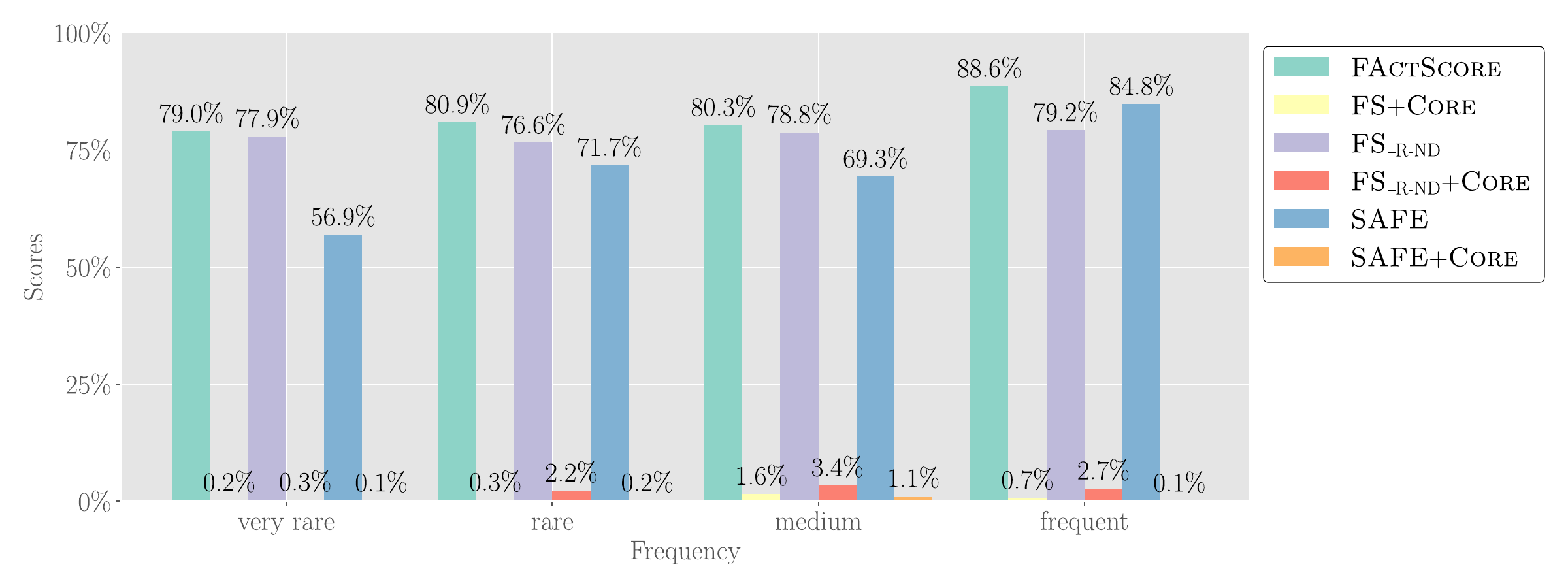}
    \caption{{\fontfamily{qcr}\selectfont freqValue} breakdowns of Factual Precision for uninformative \textbf{GPT-2}.}
\end{figure}

\begin{figure}[htpb]
    \centering
    \includegraphics[width=.95\textwidth]{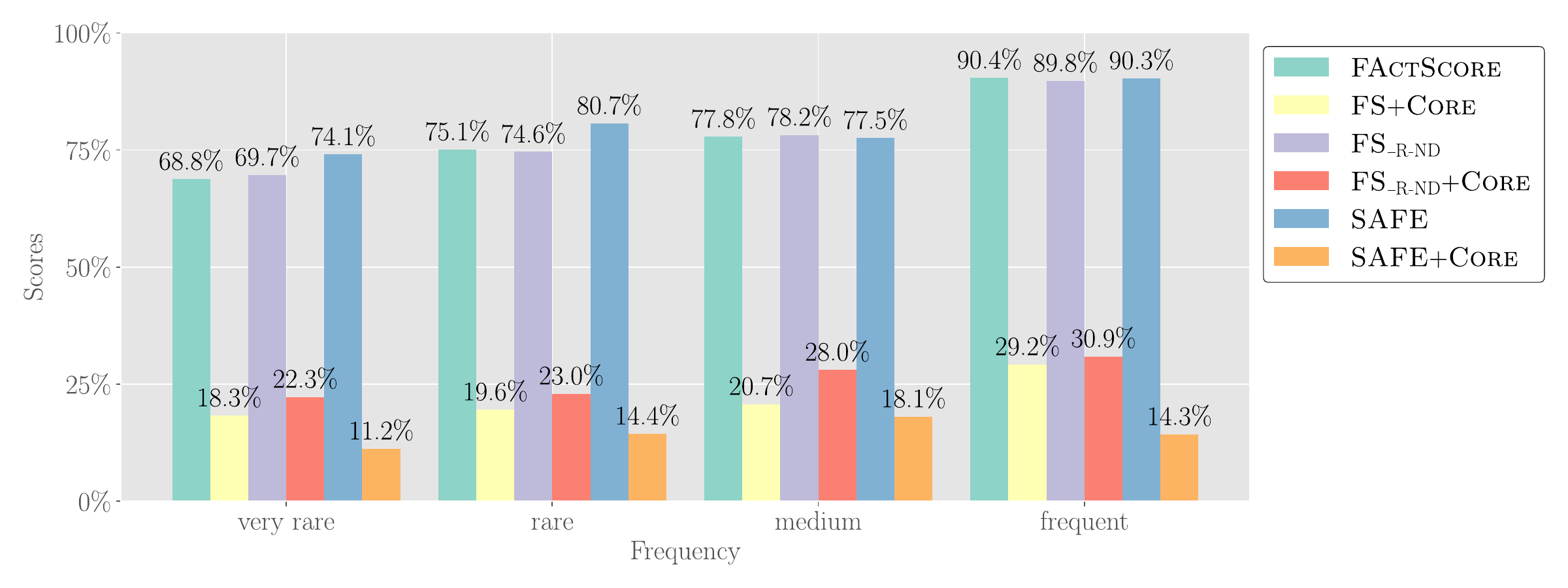}
    \caption{{\fontfamily{qcr}\selectfont freqValue} breakdowns of Factual Precision for repetitive \textbf{Mistral\textsubscript{INST}}.}
\end{figure}

\begin{figure}[htpb]
    \centering
    \includegraphics[width=.95\textwidth]{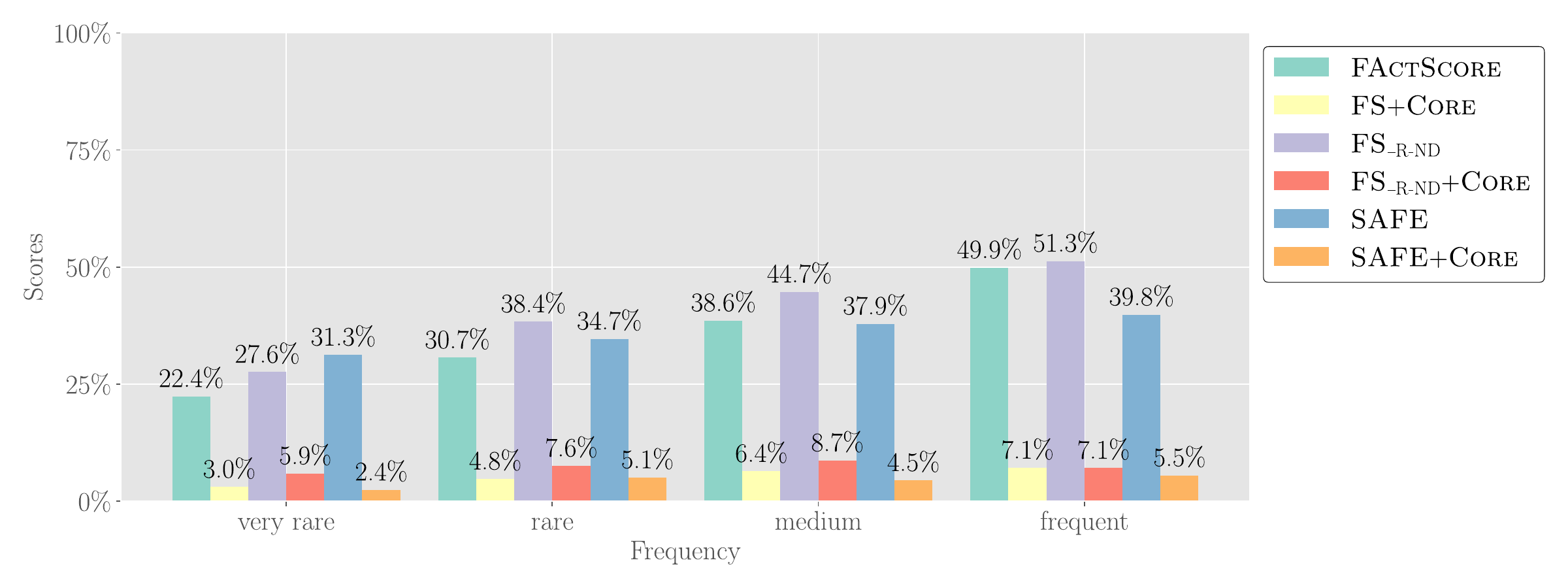}
    \caption{{\fontfamily{qcr}\selectfont freqValue} breakdowns of Factual Precision for repetitive \textbf{GPT-2}.}
\end{figure}

\end{document}